\documentclass[letterpaper, 10 pt, conference]{ieeeconf}  

\IEEEoverridecommandlockouts                              

\usepackage{graphics} 
\usepackage{amsmath} 
\usepackage{amssymb}  
\usepackage[table]{xcolor}
\usepackage[hidelinks]{hyperref}
\usepackage{subcaption}
\usepackage{graphicx}
\usepackage{wasysym}

\usepackage{tcolorbox}
\usepackage{booktabs} 
\usepackage{siunitx}
\usepackage{textcomp}
\usepackage{tabularx,ragged2e}
\newcolumntype{C}{>{\Centering\arraybackslash}X} 
\newcolumntype{P}[1]{>{\centering\arraybackslash}p{#1}} 
\usepackage{booktabs}
\usepackage{makecell}
\usepackage{multirow}

\usepackage{tikz}
\usepackage{lipsum}
\usepackage{xcolor}

\newcommand\copyrighttext{%
    \footnotesize \color{gray} 
    This work has been submitted to the IEEE for possible publication. Copyright may be transferred without notice, after which this version may no longer be accessible.
    }
\newcommand\copyrightnotice{%
\begin{tikzpicture}[remember picture,overlay]
\node[anchor=south,yshift=10pt] at (current page.south) {\parbox{\dimexpr\textwidth-\fboxsep-\fboxrule\relax}{\copyrighttext}};
\end{tikzpicture}%
}

\newcommand\scalemath[2]{\scalebox{#1}{\mbox{\ensuremath{\displaystyle #2}}}}

\title{\LARGE \bf
MoSS: Monocular Shape Sensing for Continuum Robots
}

\author{Chengnan Shentu$^{*,1,3}$,~\IEEEmembership{Student~Member,~IEEE},
        Enxu Li$^{*,1,4}$,~\IEEEmembership{Student~Member,~IEEE},\\
        Chaojun Chen$^{2}$,
        Puspita Triana Dewi$^{2,3}$,
        David B. Lindell$^{1}$,~\IEEEmembership{Member,~IEEE},\\
        and 
        Jessica Burgner-Kahrs$^{1,2,3}$,~\IEEEmembership{Senior~Member,~IEEE}
\thanks{$^*$ indicates equal contribution.}
\thanks{$^1$Department of Computer Science, University of Toronto, Canada}
\thanks{$^2$Department of Mechanical and Industrial Engineering, University of Toronto, Canada}
\thanks{$^3$Continuum Robotics Laboratory, Department of Mathematical and
Computational Sciences, University of Toronto, Canada}
\thanks{$^4$Waabi, Canada}
\thanks{We acknowledge the partial support of this research by the Natural Sciences and Engineering Research Council of Canada (NSERC), [RGPIN-2019-04846]. Corresponding author: \href{mailto:cshentu@cs.toronto.edu}{\tt\small cshentu@cs.toronto.edu}}
}

\begin{document}

\maketitle
\thispagestyle{empty}
\pagestyle{empty}

\copyrightnotice
\begin{abstract}
Continuum robots are promising candidates for interactive tasks in medical and industrial applications due to their unique shape, compliance, and miniaturization capability. 
Accurate and real-time shape sensing is essential for such tasks yet remains a challenge. 
Embedded shape sensing has high hardware complexity and cost, while vision-based methods require stereo setup and struggle to achieve real-time performance. 
This paper proposes a novel eye-to-hand monocular approach to continuum robot shape sensing. 
Utilizing a deep encoder-decoder network, our method, MoSSNet, eliminates the computation cost of stereo matching and reduces requirements on sensing hardware. 
In particular, MoSSNet comprises an encoder and three parallel decoders to uncover spatial, length, and contour information from a single RGB image, and then obtains the 3D shape through curve fitting.
A two-segment tendon-driven continuum robot is used for data collection and testing, demonstrating accurate (mean shape error of 0.91 mm, or 0.36\% of robot length) and real-time (70 fps) shape sensing on real-world data.
Additionally, the method is optimized end-to-end and does not require fiducial markers, manual segmentation, or camera calibration. 
Code and datasets will be made available at \url{https://github.com/ContinuumRoboticsLab/MoSSNet}.
\end{abstract}

\section{Introduction}

Continuum robots are robotic manipulators that do not contain rigid links or identifiable joints.
They have been studied for interactive applications such as minimally invasive surgery~\cite{burgner2015continuum} and non-destructive inspection~\cite{wang2021design}.
To enable continuum robots to perform these tasks in a flexible and adaptable manner, accurate and real-time 3D shape sensing is crucial---by tracking the robot's shape in the environment, controllers can be applied to reach a desired position, avoid collisions, or follow a certain path. Additionally, shape sensing allows the monitoring of the robot's condition and performance.

Model-based shape reconstruction is proposed to estimate the 3D shape of continuum robots from internal sensors in the drive-system and actuators, but they have a 
trade-off between accuracy and computation cost \cite{kim2015real}. 
Additionally, they are sensitive to uncertainty in the modeling parameters, unmodelled effects, and unknown external loads. 
To address these limitations, sensing-based approaches have been proposed, which can be broadly categorized as embedded or vision-based methods.
\begin{figure}[t!]
    \includegraphics[width=8.5cm]{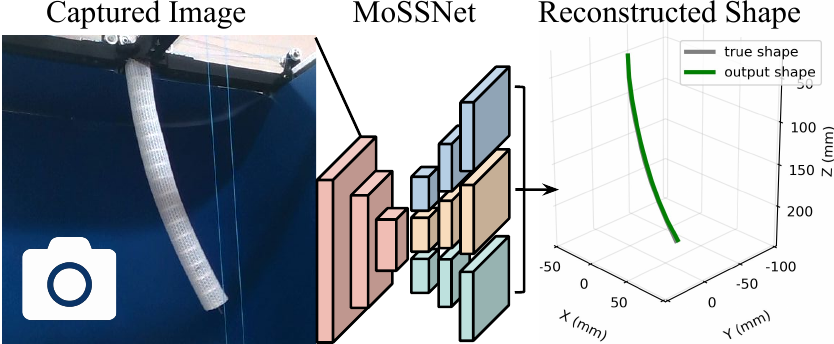}
    \caption{Our method, MoSSNet, takes a single camera image as input and outputs an accurate parametric representation of the robot centerline in real-time, without requiring fiducial markers, manual segmentation or camera calibration.}
    \label{fig:continuum}
\end{figure}

 Embedded sensors, such as fiber-optic sensors, electromagnetic (EM) sensors, and force/torque sensors, provide indirect measurements for shape reconstruction, but they require customized integration efforts and can be sensitive to external interference \cite{franz2014electromagnetic}. 
 Vision-based methods, on the other hand, offer high accuracy at low cost and can be easily adapted to different robots. However, existing methods require calibrated, multi-view camera systems, and most are not capable of real-time applications, limiting their application outside a lab environment~\cite{shi2016shape}. 

\begin{table*}[t]
  \centering
  \begin{tabular}{l P{2cm} P{2cm} P{2cm} P{2cm} r}
  \toprule
  Sensing method & Error & Update rate & \makecell{Hardware\\complexity} & \makecell{Line-of-sight\\requirement} & References \\
  \midrule
  Electromagnetic tracking  & \cellcolor{green!25}low & \cellcolor{yellow!25}medium to high & \cellcolor{red!25}high & \cellcolor{green!25}none & \cite{song2015real,condino2012electromagnetic}\\
  Optical strain sensor     & \cellcolor{green!25}very low & \cellcolor{green!25}high & \cellcolor{red!25}high & \cellcolor{green!25}none & \cite{monet2020high,Cao2022spatial}\\
  Force/torque sensor       & \cellcolor{red!25}high & \cellcolor{green!25}very high & \cellcolor{yellow!25}medium & \cellcolor{green!25}none & \cite{donat2021real}\\
  Stereo vision             & \cellcolor{green!25}low & \cellcolor{red!25}low & \cellcolor{green!25}low & \cellcolor{red!25}two & \cite{burgner_herrell_webster_2011,delmas2015three,croom_rucker_romano_webster_2010}\\
  \textbf{Monocular vision (Ours)}   & \cellcolor{green!25}low & \cellcolor{green!25}high & \cellcolor{green!25}low & \cellcolor{yellow!25}one & \\
  \bottomrule
  \end{tabular}
  \caption{Overview of methods for continuum robot shape sensing. Sensors without line of sight requirements have high hardware complexity/cost, while stereo methods tend to be slow and require line of sight to multiple cameras. Our method is low-cost, accurate, efficient, and requires line of sight to only a single camera.}
  \label{tab:sota_overview}
\end{table*}

This tradeoff between hardware complexity and performance motivates our method, MoSSNet, which is an end-to-end deep encoder-decoder network for monocular shape sensing, as illustrated in
Fig. \ref{fig:continuum}.
Our main contributions can be summarized as, 1) a novel monocular approach for continuum robot shape sensing that is accurate (mean shape error of $\SI{0.91}{\milli\meter}$ or $0.36\%$ of robot length) and real-time ($70$ fps) without requiring fiducial markers, segmentation, or camera calibration; 2) a comprehensive evaluation of its quantitative and qualitative performance in simulation and on hardware ; 3) a public dataset for future development and benchmarking of monocular continuum robot shape sensing.

\section{Related Work}
\label{sec:related}

In this section, we discuss the diverse approaches taken for 3D shape sensing of continuum robots. Based on the sensing principle, we categorize them into embedded and vision-based methods and discuss them separately. A brief overview of these approaches is presented in Table \ref{tab:sota_overview}. 

\subsection{Embedded Shape Sensing}

Magnetic sensors, typically used in electromagnetic (EM) tracking systems, can be attached to the robots and localized by measuring the small current induced by a magnetic field generated in the workspace. Thus, such methods are sensitive to EM interference and have limited workspace, which poses constraints on the robot material and application environments \cite{franz2014electromagnetic}. 
Because each sensor only provides pose information at a single point, determining the number of sensors to use is crucial. A higher number of sensors can lead to increased accuracy, but rigid sensors can interfere with the continuum robot's characteristics. Conversely, a lower number of sensors are more efficient, but rely more heavily on kinematic models \cite{condino2012electromagnetic} or shape representations \cite{song2015real}, which can make the process more computationally expensive and vulnerable to model uncertainties.

Strain sensors, such as optical fibers with fiber Bragg gratings (FBGs), measure shape by integrating local curvature/strain. Their small size and bio-compatibility allow embedded shape sensing in varied environments with real-time capability and has motivated research and commercialization efforts~\cite{monet2020high,Cao2022spatial}. 
However, they require customized sensor setups for different robots, which leads to high costs \cite{da2020challenges}. 

Lower-cost alternatives such as passive strings \cite{li2022shape} have been proposed at the expense of lower accuracy and update rate. Orekhov et al.~\cite{orekhov2023lie} formally analyzed the problem of string routing optimization through sensitivity analysis, and achieved tip error of $\SI{5.9}{\milli\meter} (1.97\%)$ at $4$ fps. Moreover, they share the same drawback of being highly sensitive to the placement and calibration of the sensors.

In addition to electromagnetic tracking and strain-based shape sensing, force/torque sensors have been employed to estimate the shape of elastic rods using the Kirchhoff rod model \cite{nakagawa2018real}. This method is cost-effective and computationally efficient. While attempts have been made to extend it to multi-segment continuum robots, position errors increase with larger deformation due to unmodeled effects \cite{donat2021real}.

Common to all embedded shape sensing approaches is they do not require line-of-sight, which is valuable for applications in confined spaces.
However, they have a higher level of hardware complexity because they require customized sensor integration and calibration efforts. Their accuracy and update rate scale with the performance of sensing hardware, which also leads to a high cost of deployment (see Table~\ref{tab:sota_overview}).

\subsection{Vision-based Shape Sensing}

\begin{figure*}[t]
    \centering
    \includegraphics[width=7in]{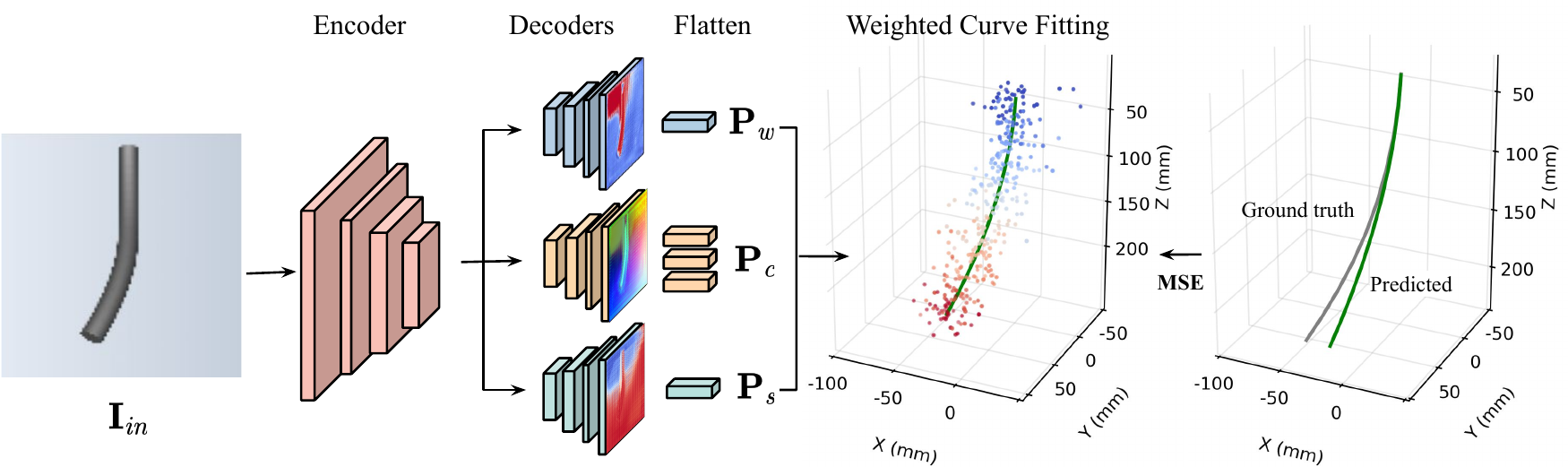}
    \caption{Overview of our approach, MoSSNet. The network takes as input the captured image of the robot and generates importance for reconstruction, centerline coordinates, and relative arclength. These flattened outputs are then processed by the weighted curve fitting algorithm to generate a curve that parameterizes the robot's centerline. To train the network, we supervise its learning process by penalizing the mean squared error between the predicted and ground truth curves.}
    \label{fig:overview}
\end{figure*}

Vision-based eye-to-hand shape sensing has gained attention since they are cost effective and have low hardware complexity. One of the most straightforward approaches is detecting point correspondence in stereo setups. Delmas et al. achieved mean error of $\SI{0.46}{\milli\meter}~(2.30\%)$ at $6.25$ fps in simulated fluoroscopic image pairs \cite{delmas2015three}. Such methods can be slow for continuum robots, because they lack identifiable joints and links as feature points. 

Fiducial markers can be added to speed up point correspondence \cite{li2020marker}, but integrating markers that meet size, shape, and visibility constraints is difficult in applications. Another approach to speed up stereo shape sensing is to apply simplifying assumptions, such as limiting the shapes to quadratic forms, as done by Dalvand et al.~\cite{dalvand_nahavandi_howe_2016} However, this approach does not generalize to robots capable of more complex shapes.
Camarillo et al.~\cite{camarillo2008vision} proposed 3D shape reconstruction using the shape-from-silhouette technique with three cameras. Self-organizing maps algorithm is also investigated by Croom et al.~\cite{croom_rucker_romano_webster_2010} to fit a 3D curve from point cloud.
Although these methods are accurate, and achieve mean error of $\SI{1.14}{\milli\meter} (0.72\%)$ and $\SI{1.53}{\milli\meter} (0.64\%)$ respectively, they still require point correspondence and are computationally expensive ($\leq 4$ fps). 

At its core, vision-based shape sensing requires some depth information on the robot to reconstruct its 3D shape, which is traditionally achieved through epipolar geometry with two or more cameras. 
Inspired by recent advances in monocular depth estimation \cite{alhashim2018high} and monocular 3D object reconstruction \cite{zhou2021monocular}, 
which utilizes deep learning architectures to tackle these inherently ill-posed problems, 
we propose a novel eye-to-hand monocular shape sensing approach for continuum robots with an encoder-decoder network, thus eliminating the computation cost of stereo matching and reducing requirements on sensing hardware.
Our approach, MoSSNet, achieves accurate (mean shape error of $\SI{0.91}{\milli\meter} (0.36\%)$) and real-time ($70$ fps) results on a real-world dataset. The method takes a single RGB image as input and outputs a parametric representation of the robot centerline, without requiring fiducial markers, manual segmentation, or camera calibration. 

\section{MoSSNet}
\label{sec:method}

In this section, we introduce Monocular Shape Sensing Network (MoSSNet), an efficient and effective approach for continuum robot shape estimation. The problem formulation, network architecture, and training methodology are introduced in the following.

\subsection{Problem Formulation}

We suppose there is a camera that remains in a fixed pose relative to the robot's base. The camera captures an image of the robot, which we refer to as $\mathbf{I}_{RGB} \in \mathbb{R}^{H \times W \times 3}$ where $H$ and $W$ are the height and width of the captured image. The objective is to determine the 3D centerline of the robot, which is the widely adopted representation of continuum robots' shape \cite{shi2016shape}. Specifically, we identify the coordinates of $M$ equally spaced points along its centerline, represented as $\mathbf{P}_{r} \in \mathbb{R}^{M \times 3}$.

In this work, we limit the scope to monocular RGB images with uniform lighting conditions and no occlusions except the robot's self-occlusions. We also assume the robot's geometry remains constant and its motion generates minimal to no motion blur on the images captured. 

\subsection{Network Architecture}
In order to reconstruct the robot, our model uses an image of the robot to predict the 3D coordinates along its centerline. 
Subsequently, a weighted linear least squares algorithm is employed to derive a 3D curve that parametrizes the center of the robot. 
As shown in Fig. \ref{fig:overview}, we design a network with a shared encoder and three decoders that are composed of four stages each. These stages consist of a residual block \cite{resnet} with two convolutional layers that are connected through BatchNorm \cite{ioffe2015batch} and Leaky ReLU activations \cite{maas2013rectifier}. The encoder block uses a maxpooling layer between every two stages to decrease the feature map's size by a factor of two. In contrast, the decoder block incorporates a pixel shuffle layer \cite{shi2016shuffle} between every two stages to increase the feature map's size by a factor of two.
Next, we explain each component in more detail.

\paragraph{Encoder}
To incorporate location information, we add the 2D image indices to  $\mathbf{I}_{RGB}$. This results in a 5-channel image, represented as $\mathbf{I}_{in} \in \mathbb{R}^{H \times W \times 5}$. The encoder then extracts multi-scale features from the input image, which are subsequently passed through three decoders for further processing.

\paragraph{Decoders}
Given that the image includes background, not every pixel is relevant in determining the robot's shape. To address this, the importance decoder learns the significance of each pixel in shape reconstruction. The output of the importance decoder is a per-pixel importance score denoted as $\mathbf{P}_w \in \mathbb{R}^{HW}$, which is normalized between $0$ and $1$ using a Sigmoid function. A higher value of this score indicates that the pixel is more relevant for shape sensing, whereas a lower value suggests that it belongs to the background. These scores will be used to perform weighted curve fitting in later stages.

Subsequently, we use another decoder to generate an estimate of the robot's shape by predicting its $xyz$ coordinates along its centerline, denoted as $\mathbf{P}_c \in \mathbb{R}^{HW \times 3}$.
It is worth noting that only regions that cover the robot's shape, i.e., regions with high importance scores, will have useful values. 
Further, the last decoder learns a per-pixel relative arclength that ranges from 0 to 1, where 0 represents the robot's base and 1 represents its tip, denoted by $\mathbf{P}_s \in \mathbb{R}^{HW}$. 

\begin{figure*}[ht!]
   \subfloat[\label{prototype}]{%
      \includegraphics[width=9.24cm]{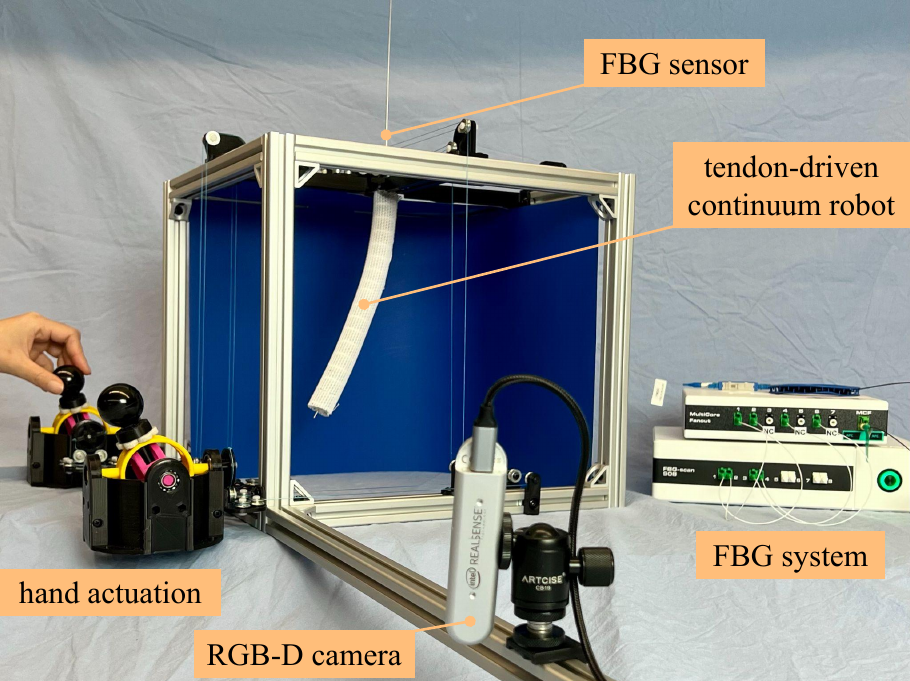}}
\hspace{\fill}
   \subfloat[\label{real_image} ]{%
      \includegraphics[width=3.9cm]{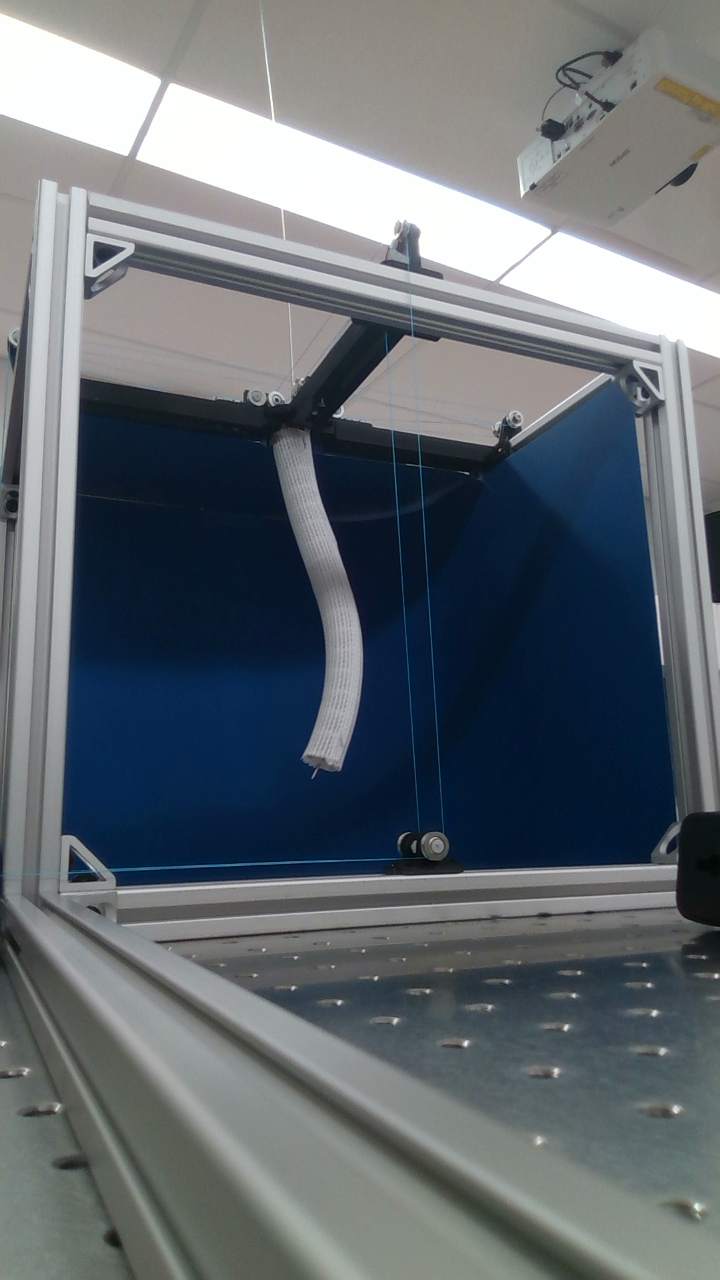}}
\hspace{\fill}
   \subfloat[\label{sim_image}]{%
      \includegraphics[width=3.9cm]{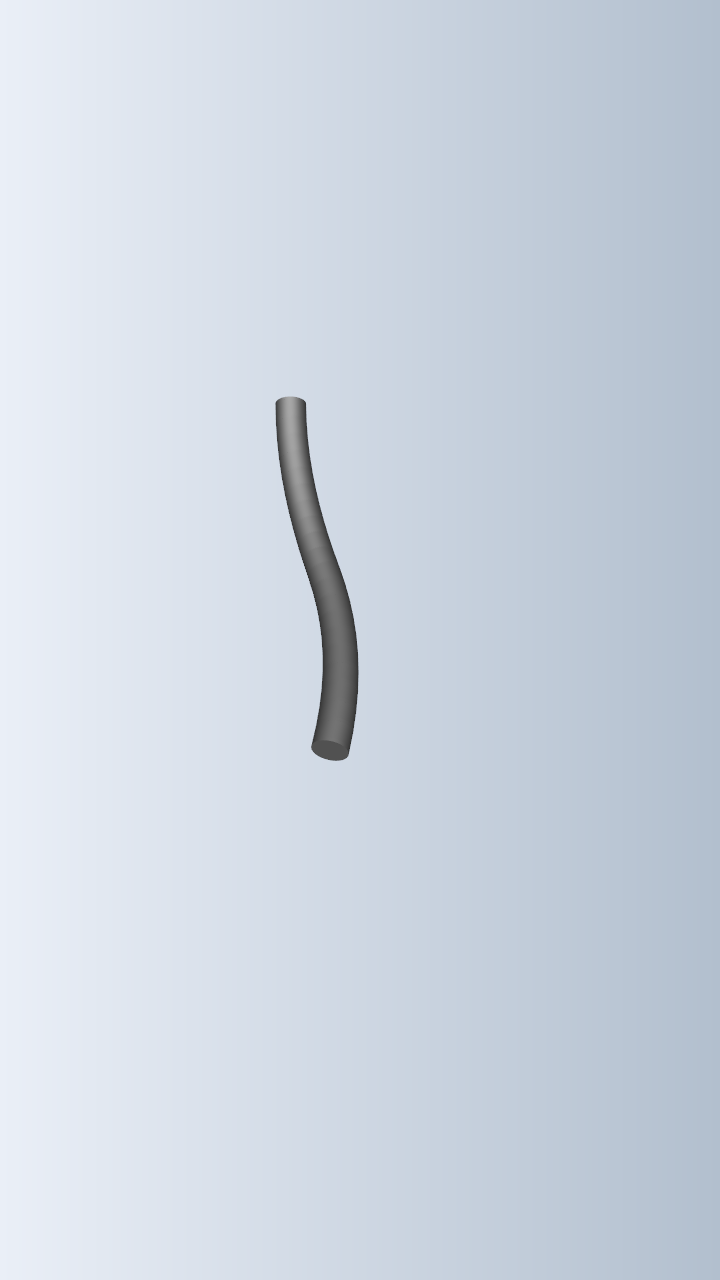}}\\
\caption{We collect a monocular shape sensing dataset with a two-segment tendon-driven continuum robot on hardware and in simulation (a) overview of hardware setup; (b) sample captured image; (c) sample simulated image.}
\end{figure*}

\paragraph{Weighted Curve Fitting}
We model the robot centerline as three independent $n$-th order polynomials on $x$, $y$, and $z$ axes with curve parameters being $\mathbf{w} = [\mathbf{w}_x \ \mathbf{w}_y \ \mathbf{w}_z] \in \mathbb{R}^{(n+1) \times 3}$. Formally, we have
\begin{equation}
\scalemath{0.9}{
    \underbrace{
    \begin{bmatrix}
    \vert & \vert & \vert & \vert\\
    (\mathbf{P}_s)^0   & (\mathbf{P}_s)^1 & ... & (\mathbf{P}_s)^n   \\
    \vert & \vert & \vert & \vert
\end{bmatrix}
}_{\mathbf{A} \in \mathbb{R}^{HW \times (n+1)}}
\underbrace{
\begin{bmatrix}
    \vert & \vert & \vert\\
    \mathbf{w}_x   & \mathbf{w}_y & \mathbf{w}_z   \\
    \vert & \vert & \vert
\end{bmatrix}
}_{\mathbf{w} \in \mathbb{R}^{(n+1) \times 3}}
= \underbrace{
\mathbf{P}_c}_{\mathbf{B} \in \mathbb{R}^{HW \times 3}}}.
\end{equation}
To obtain a curve of best fit, we compute the weighted least squares solution of the curve parameter $\mathbf{w}$ as follows,
\begin{equation}
    \mathbf{w} = (\mathbf{A}^T \mathbf{\Sigma} \mathbf{A})^{-1} \mathbf{A}^T \mathbf{\Sigma} \mathbf{B},
\end{equation}
where $\mathbf{\Sigma} = \text{diag}(\mathbf{P}_w) \in \mathbb{R}^{HW \times HW}$ is the learned per-pixel weighting for the curve fitting.
Finally, we obtain the coordinates of the M evenly-spaced points along the robot centerline by querying M relative locations.
\begin{equation}
\scalemath{0.9}{
    \underbrace{
    \begin{bmatrix}
        (1/M)^0 & (1/M)^1 & ... & (1/M)^n \\
        (2/M)^0 & (2/M)^1 & ... & (2/M)^n \\
        \vdots \\
        (M/M)^0 & (M/M)^1 & ... & (M/M)^n
    \end{bmatrix}}_{\mathbf{P}_q \in \mathbb{R}^{M \times (n+1)} }\mathbf{w} =
    \underbrace{
    \begin{bmatrix}
    \vert & \vert & \vert\\
    \hat{\mathbf{p}}_x & \hat{\mathbf{p}}_y & \hat{\mathbf{p}}_z  \\
    \vert & \vert & \vert
    \end{bmatrix}
    }_{\hat{\mathbf{P}}_r \in \mathbb{R}^{M \times 3} }
    }.
\end{equation}
\subsection{Supervision}
We use the weighted mean squared error of the $M$ predicted coordinates and the ground truth as the loss function to supervise the network.
The loss function is depicted as follows,
\begin{equation}
    L =  \frac{1}{M} \sum_{j=1}^{M} \beta_j \lVert \hat{\mathbf{P}}_{r,j} - \mathbf{P}_{r,j} \rVert_{2}^{2},
\end{equation}
where $\beta_j$ is the weight applied on each of the $M$ points. Given the inherent difficulty in accurately reconstructing the tip location of the robot, we assign a larger weight on $\beta_M$ to penalize the tip error.

\section{Data Collection and Benchmark}
\label{sec:dataset}

The monocular shape sensing method proposed is evaluated both in simulation and on a robot prototype. While the method can be applied to all types of continuum robots, we focus on tendon-driven continuum robots (TDCRs), which are one of the most used and researched continuum robot types. 

First, we describe the hardware and simulation setup for data collection. Afterward, we provide an overview of our dataset and define the metrics used for benchmarking.

\subsection{Hardware Setup}
For this work we use a TDCR that is $\SI{250}{\milli\meter}$ in length and $\SI{20}{\milli\meter}$ in diameter. 
It has two identical bending segments with 20 equally distributed spacer disks for tendon routing along a super-elastic Nitinol backbone. 
The robot is covered in a flexible polyethylene sleeve.
The TDCR is mounted on a mechanical frame and tendons are manually operated.

We collect RGB and depth images using an RGB-D camera (RealSense D415, Intel, USA) at $1280 \times 720$ resolution. The ground truth shape of the TDCR is measured by an FBG shape sensing system (custom multicore fiber, fan-out box, and FBG-Scan 908 interrogator, FBGS International NV, Belgium) placed inside the backbone. The multicore fiber sensor has sensing length of $\SI{250}{\milli\meter}$, which contains $26$ evenly spaced gratings, and outputs a shape measurement of $251$ points. 
The hardware setup is shown in Fig. \ref{prototype}, and a sample RGB image is shown in Fig. \ref{real_image}. 
During data collection, the camera allows simultaneous RGB and depth image capture at 6Hz, while the FBG system provides updates at 100Hz. We collect the shape measurement with the closest timestamp for each camera frame, resulting in a worst-case time difference of $\SI{5}{\milli\second}$.

The FBG sensor is positioned such that its coordinate frame aligns with that of the robot. 
During training, the RGB images are inputs to the network. Points along the robot centerline are given directly by the FBG system and are used as ground truth. Depth images are not used in our method, but are included in the dataset to facilitate further research.

\subsection{Simulation Setup}
\label{subsec:dataset_sim}
A TDCR with the same parameters is simulated using the Cosserat rod-based static model \cite{rucker2011statics}, specifically a C++ implementation published in \cite{rao2021model}.
The simulated robot is then rendered using the Visualization Toolkit (VTK) with simple texture and lighting \cite{schroeder1998visualization}. 
We apply calibrated camera intrinsic and extrinsic from the hardware setup to obtain the same view, as shown by the sample image in Fig. \ref{sim_image}. By sampling random joint space configurations, we simulate realistic robot shapes and obtain RGB and depth images from VTK. The model implementation also allows us to sample robot shapes under external load, so we also simulate robot shapes with random external force and moment applied at the tip. 

For training, the RGB images are inputs to the network. Points along the robot's centerline, provided by the simulator, are used as the ground truth.

\begin{figure*}[ht!]
\centering
\begin{tabular}{c|cccc|c}
\includegraphics[width=2.6cm,height=2.6cm]{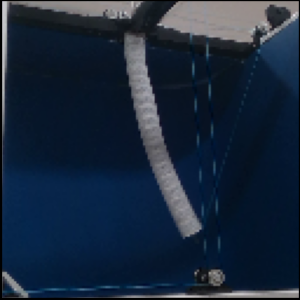} & \includegraphics[width=2.6cm,height=2.6cm]{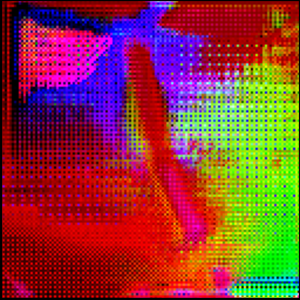} & 
\includegraphics[width=2.6cm,height=2.6cm]{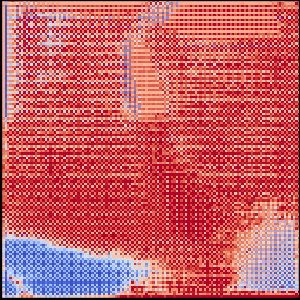} & 
\includegraphics[width=2.6cm,height=2.6cm]{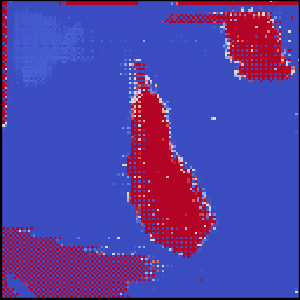} & 
\multirow{2}{*}[1cm]{\includegraphics[height=3cm]{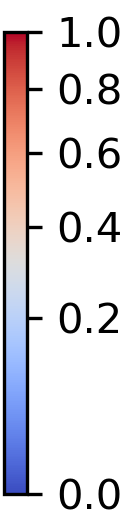}}
 &
\includegraphics[height=2.8cm]{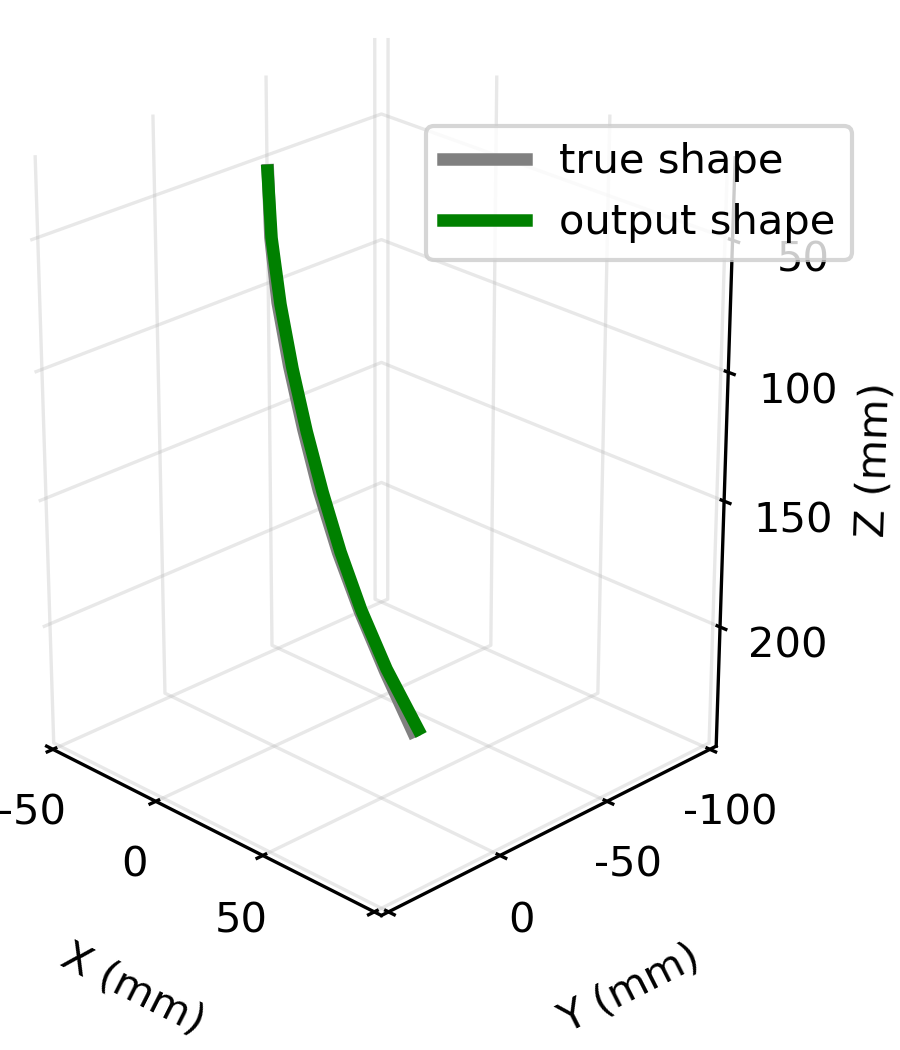}\\
\includegraphics[width=2.6cm,height=2.6cm]{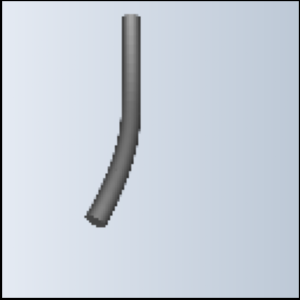} & \includegraphics[width=2.6cm,height=2.6cm]{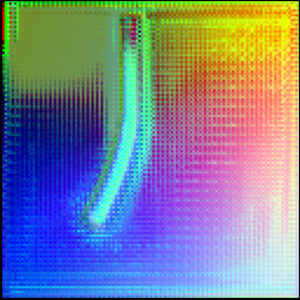} & 
\includegraphics[width=2.6cm,height=2.6cm]{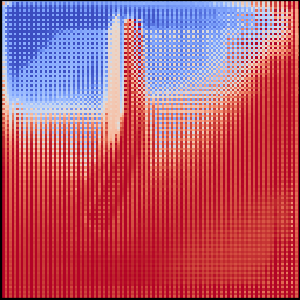} & 
\includegraphics[width=2.6cm,height=2.6cm]{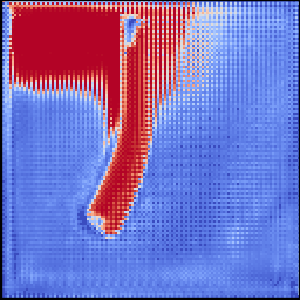} & 
&
\includegraphics[height=2.8cm]{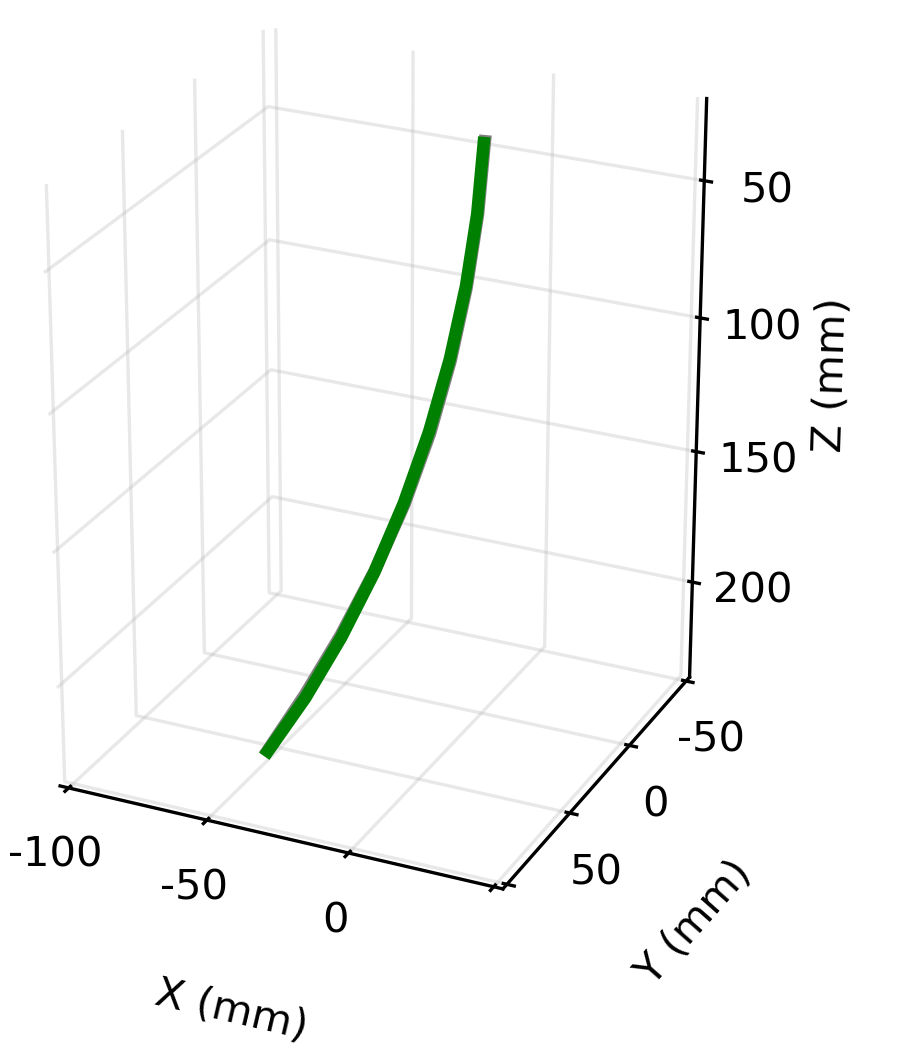}\\
Input image & Centerline decoder & Arclength decoder & Importance decoder && Output shape
\end{tabular}
\caption{\label{fig:quali}MoSSNet's three decoders output interpretable pixel-wise information for 3D curve fitting despite not having pixel-wise supervision. The three encoder outputs, from left to right, provide spatial, length, and contour information about the robot to obtain accurate shape via weighted curve fitting.
}
\end{figure*}

\subsection{MoSS-Real and MoSS-Sim Datasets}
We present our dataset for monocular shape sensing of a two-segment TDCR. Using the hardware and simulation setup outlined above, we collect $12\,000$ configurations on the robot prototype and $50\,000$ configurations rendered in the simulator, including $25\,000$ in free space and $25\,000$ with external load at the tip with 3 force components and 3 moment components. Each force component is randomly sampled between $[\SI{-0.1}{\newton}, \SI{0.1}{\newton}]$, and each moment component is randomly sampled between $[\SI{-0.01}{\newton\meter}, \SI{0.01}{\newton\meter}].$ 
We will refer to the two datasets as the MoSS-Real dataset and the MoSS-Sim dataset in later sections of this paper.
We split both datasets into training, validation, and test sets randomly with ratios of $60\%$, $15\%$, and $25\%$ respectively.
Depth images and camera parameters are not used in our proposed method but are included to support the development of alternative methods.

\subsection{Benchmarking metrics}

Shape sensing for continuum robots is typically evaluated in terms of mean error of robot shape (MERS) and mean error of robot tip (MERT) \cite{shi2016shape}. 
Although they have been calculated differently across literature, we define MERS to be the average Euclidean distance between the predicted set of evenly-spaced points, $\hat{\mathbf{P}}_{r} \in \mathbb{R}^{M \times 3}$, and corresponding ground truth points,  $\mathbf{P}_{r} \in \mathbb{R}^{M \times 3}$, across different configurations in the robot's workspace. 

\begin{equation}
    \text{MERS} = \frac{1}{M} \sum_{j=1}^{M} \lVert \hat{\mathbf{P}}_{r,j} - \mathbf{P}_{r,j} \rVert_{2}.
\end{equation}

Shape sensing output should provide information dense enough to reconstruct the complete robot shape for applications like collision checking.
Thus, the minimum number of output points depend on the complexity of robot shape representation (i.e., degree-of-freedom in configuration space), such that there are enough data points for model fitting. 
We further constraint $M$ to be at least double the minimum number of points to take account for errors from robot shape representation and avoid aliasing.
In our case, $4^\text{th}$ order polynomial is used and requires 5 points for curve fitting, so we constraint $M \geq 10$.
MERT is calculated in the same way but only accounting for the tip position.
\begin{equation}
    \text{MERT} = \lVert \hat{\mathbf{P}}_{r,M} - \mathbf{P}_{r,M} \rVert_{2}.
\end{equation}

To facilitate comparison of results between different robots, we also report MERS and MERT with respect to the sensed length of the robot as a percentage. In our case the entire TDCR is being sensed, so we divide the error by $\SI{250}{\mm}$.
We also evaluate our method's real-time capability by reporting its update rate in frames per second (fps). 
We evaluate our method against these three metrics on a large number of different shapes to ensure its robustness. 

\section{Evaluation}

In this section, we first present the implementation details and then show the quantitative and qualitative results of the proposed approach on MoSS-Sim and MoSS-Real datasets. Further, we explore the approach's sim-to-real transfer capability and provide an ablation analysis on various components of the network. Finally, we discuss the robustness of our approach when the camera configuration changes.

\subsection{Implementation Details}
To process the captured image, we first crop it to $512 \times 512$ using a manually defined region of interest. We then use nearest downsampling to scale the image to $128 \times 128$ before feeding it into our network. We use a fourth order polynomial to model the centerline of the robot, and take $M=10$. During training, we use a batch size of $4$ and run our experiments on an NVIDIA T4 GPU. For both simulated and real datasets, we train our network with the AdamW optimizer using a constant learning rate of $0.001$ for $150$ epochs. The loss weight $\beta_j$ is set to $1$ for $j = 1 ... M-1$ and $2$ for $j=M$.
We evaluate our method's update rate on the test set, with a batch size of 1 to simulate sequential input image data, on the same computer (GPU: NVIDIA GeForce RTX 3090, CPU: AMD - Ryzen 5950x 3.4 GHz 16-core 32-thread, RAM: 64GB, OS: Ubuntu 20.04). Specifically, we measure the time between input data being passed to the network and shape sensing results being received. This includes not only the inference time but also data transfers between CPU and GPU, thus is more reflective of real-life update rate.

\subsection{Quantitative Results}
We report the shape sensing accuracy and runtime of our method, which are trained and tested on MoSS-Real and MoSS-Sim datasets separately. Additionally, MoSS-Real does not contain external effects besides gravity and friction from the sleeve. To explore if the method can generalize to robot shapes unseen during training, we collect an additional test set of $5\,000$ shapes on hardware with external forces, achieved by attaching a $\SI{20}{\gram}$ calibrated weight to the robot tip. The weight is attached with a blue polyethylene string and hung outside the mechanical frame to minimize the influence on the captured images. We refer to this test set as Disturbed-Real.

\begin{table}[ht]
  \centering
  \begin{tabular}{lccc}
  \toprule
  Test set &
  MERS (mm) $\downarrow$&
  MERT (mm) $\downarrow$ &
  fps $\uparrow$ \\
  \midrule
  MoSS-Sim & 0.51 (0.20\%) & 0.88 (0.35\%) & 71.0\\
  MoSS-Real & 0.91 (0.36\%) & 1.85 (0.74\%) & 70.3\\
  Disturbed-Real & 1.98 (0.79\%) & 4.40 (1.76\%) & 72.0\\
    \bottomrule
  \end{tabular}
  \caption{Quantitative performance of MoSSNet on MoSS-Sim, MoSS-Real, and an additional real test set Disturbed-Real with unseen external disturbance.}
  \label{tab:quant}
\end{table}
Table \ref{tab:quant} presents the metrics obtained from our evaluation. Specifically, MoSSNet achieves a mean error of robot shape (MERS) of $\SI{0.51}{\milli\meter}~(0.20\%)$  and a mean tip error (MERT) of $\SI{0.88}{\milli\meter}~(0.35\%)$  on the simulated test set. On the real test set, MoSSNet achieves a MERS of $\SI{0.91}{\mm}~(0.36\%)$  and a MERT of $\SI{1.85}{\mm}~(0.74\%)$. 
While our approach performs well on both datasets, it appears to perform slightly better on the simulated dataset due to the ideal conditions in that environment and the larger size of the dataset. The performance in the real-world setting is hindered slightly by the presence of noise in the captured images and sensor readings. Moreover, we observed that MoSSNet generalizes well on the disturbed test set, which is the real dataset with external forces that were not seen during training. The network achieves a MERS of $\SI{1.98}{\mm}~(0.79\%)$ and a MERT of $\SI{4.40}{\mm}~(1.76\%)$ mm on this test set. The model also achieves update rates over 70 fps consistently, which makes it suitable for real-time and even dynamic applications.

\subsection{Qualitative Results}

The network's performance can also be seen in Fig. \ref{fig:quali}, where test samples from MoSS-Real and MoSS-Sim are shown along with decoder outputs and shape output. All three decoder outputs provide interpretable pixel-wise information: 1) the centerline decoder predicts a 3D coordinate for each pixel in the image; 2) the arclength decoder predicts the pixel's relative position along the robot, and we observe increasing weight toward the bottom of the image as the robot points downward, with low weights at unreachable pixels; 3) the importance decoder outputs a nearly-binary segmentation of the robot, with noisy patches located at unreachable areas. 
The decoder outputs are then flattened and used for weighted linear least squares to obtain the robot shape.
Compared to MoSS-Sim, the outputs from MoSS-Real are noisier due to more complex background and lighting condition, which aligns with our previous observation of higher shape sensing error on the real dataset. 

The decoder outputs are noisy in general because only the ground truth shape is provided to the network during training, resulting in a lack of direct supervision on each pixel. This is a design decision made for practicality considerations, since per-pixel ground truth information is difficult to obtain on real continuum robots, especially if they are small in size. In our datasets, we include depth images to facilitate development in alternative approaches.

\subsection{Sim-to-real transfer learning}

\begin{figure}[h!] 
    \includegraphics[width=9.16cm]{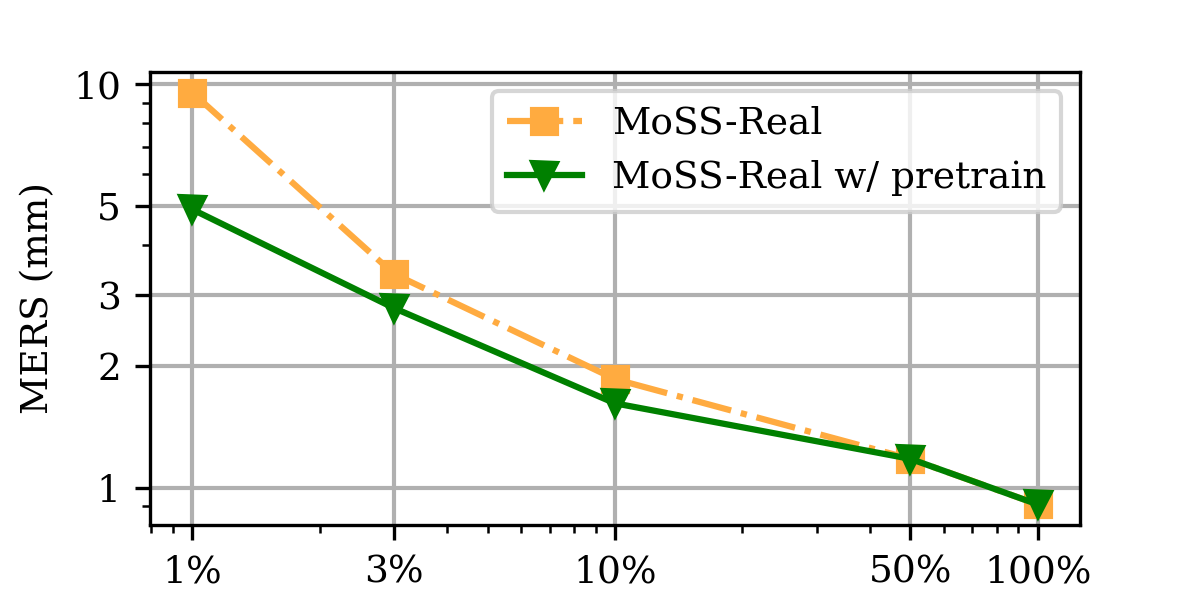}
\caption{\label{fig:sim-to-real}Influence of the amount of real training data and pre-training with simulated data on MERS.}
\end{figure}

We conduct further experiments to evaluate transfer learning from the simulated to the real dataset. Specifically, we compare the performance of training the network from scratch to that of using a network pre-trained on the simulated dataset for the real test set. We run these experiments in various settings where the amount of available training data varied. Our results are presented in Fig. \ref{fig:sim-to-real}.

Generally, a larger amount of training data leads to lower test error. 
When only a small fraction of the real training dataset was available (e.g. $1 \%$ to $10 \%$), pretraining the network on the simulated dataset leads to a significant improvement in performance on the real test set. However, when the amount of training data was large (e.g. $50\%$ and $100\%$ of the available data), pretraining on the simulated dataset does not provide additional benefit to the network's performance on the real dataset. 

Thus, pre-training the network on simulation is beneficial when data collection on real hardware is not possible or costly. It is also worth noting that the images from MoSS-Sim are very abstract and do not closely represent the setup used for MoSS-Real, so having more realistic simulated images could improve the performance gain of pretraining.

\subsection{Ablations}

\paragraph{Influence of multiple decoders}
We performed ablation studies to evaluate the contribution of each decoder, and the results are presented in Table \ref{tab:Ablation}. In the first row, we use only one decoder to generate centerline coordinates and rely on the norm of the predicted 3D coordinates as a proxy for arclength during curve fitting.
The addition of the arclength decoder enables us to predict the relative arclength accurately at each location within the captured image, resulting in a decrease of MERS by $0.13$ mm and MERT by $0.29$ mm. It's important to note that the arclength decoder has a more significant impact on reducing MERT, as it allows for better localization of the tip location. Furthermore, since not every pixel plays an equal role in regressing the robot shape, the introduction of the importance decoder assigns different weights for curve fitting to each pixel, leading to a further reduction in MERS to $0.91$ mm and MERT to $1.85$ mm.

\begin{table}[h]
\begin{center}
{
\begin{tabular}{ cccccc  }
\toprule 
\multicolumn{3}{c}{\textbf{Decoders}} & 
\multicolumn{3}{c}{\textbf{Metrics (mm or fps)}} \\ \cmidrule(lr){1-3} \cmidrule(lr){4-6}
\multicolumn{1}{c}{Centerline} &
\multicolumn{1}{c}{Arclength} &
\multicolumn{1}{c}{Importance} &
\multicolumn{1}{c}{MERS $\downarrow$} &
\multicolumn{1}{c}{MERT $\downarrow$} &
\multicolumn{1}{c}{fps $\uparrow$} \\
 \midrule 
\checkmark & & & 1.12 & 2.22 & 72.2\\
\checkmark & \checkmark & & 0.99 & 1.93 & 71.6\\
\checkmark & \checkmark & \checkmark & 0.91 & 1.85 & 70.3\\

\bottomrule
\end{tabular}
}
\end{center}
\caption{Ablation study of the proposed components.
}
\label{tab:Ablation}
\end{table}

\paragraph{Influence of polynomial degree for fitting}

We performed an ablation analysis on the polynomial degree used for representing the robot shape. From the results summarized in Table \ref{tab:polydeg}, our chosen representation of degree 4 polynomials has the lowest MERS, while degree 3 yields very similar results and has a lower MERT. We think this design parameter is dependent on the robot -- a more complex shape naturally requires higher-order representations, and that is why we see higher errors for degree 2 polynomial representation. 

Interestingly, the errors increase for degree 5 polynomial, which indicates worse representation of the achievable robot shapes. It may also be beneficial to consider other basis functions for shape representation, such as Euler curves and Chebyshev polynomials; however, formal comparison between basis functions is outside the scope of this work.

\begin{table}[h]
\begin{center}
\scalebox{0.9}
{
\begin{tabular}{ ccc  }
\toprule 

Polynomial Degree &
\multicolumn{1}{c}{MERS (mm)} &
\multicolumn{1}{c}{MERT (mm)}  \\
\midrule
2 & $2.06$ & $2.22$  \\  
3 & $0.96$ & $\mathbf{1.82}$  \\  
4 & $\mathbf{0.91}$ & $1.85$  \\  
5 & $1.01$ & $2.22$  \\    
\bottomrule
\end{tabular}
}
\end{center}
\caption{Ablation analysis on polynomial degree.}
\label{tab:polydeg}
\end{table}

\subsection{Robustness to camera configuration}
To verify that our method is robust to different imaging systems, we collected a new dataset on the simulation setup described in Section \ref{subsec:dataset_sim}. The dataset contains $20000$ shapes captured in $512 \times 512$ resolution with the same $60\%-15\%-25\%$ split. The camera is placed $\SI{50}{\milli\meter}$ from the robot's base and has a view angle of $\SI{120}{\degree}$ to simulate small wide-angle cameras used for endoscopy or industrial inspection, and we refer to this dataset as MoSS-SimWide.

\begin{table}[h!]
  \centering
  \begin{tabular}{lccc}
  \toprule
  Test set &
  MERS (mm) $\downarrow$&
  MERT (mm) $\downarrow$ &
  fps $\uparrow$ \\
  \midrule
  MoSS-SimWide & 0.78 (0.32\%) & 1.61 (0.64\%) & 103\\
    \bottomrule
  \end{tabular}
  \caption{Quantitative results on MoSS-SimWide}
  \label{tab:quant_app}
\end{table}
\begin{figure} [h!]
\centering
\begin{tabular}{cccc}
\includegraphics[width=.375\linewidth,height=.4\linewidth]{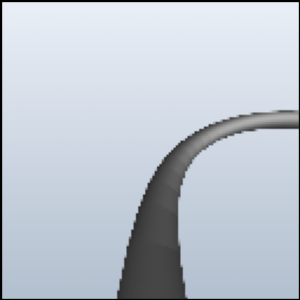} &
\includegraphics[width=.6\linewidth]{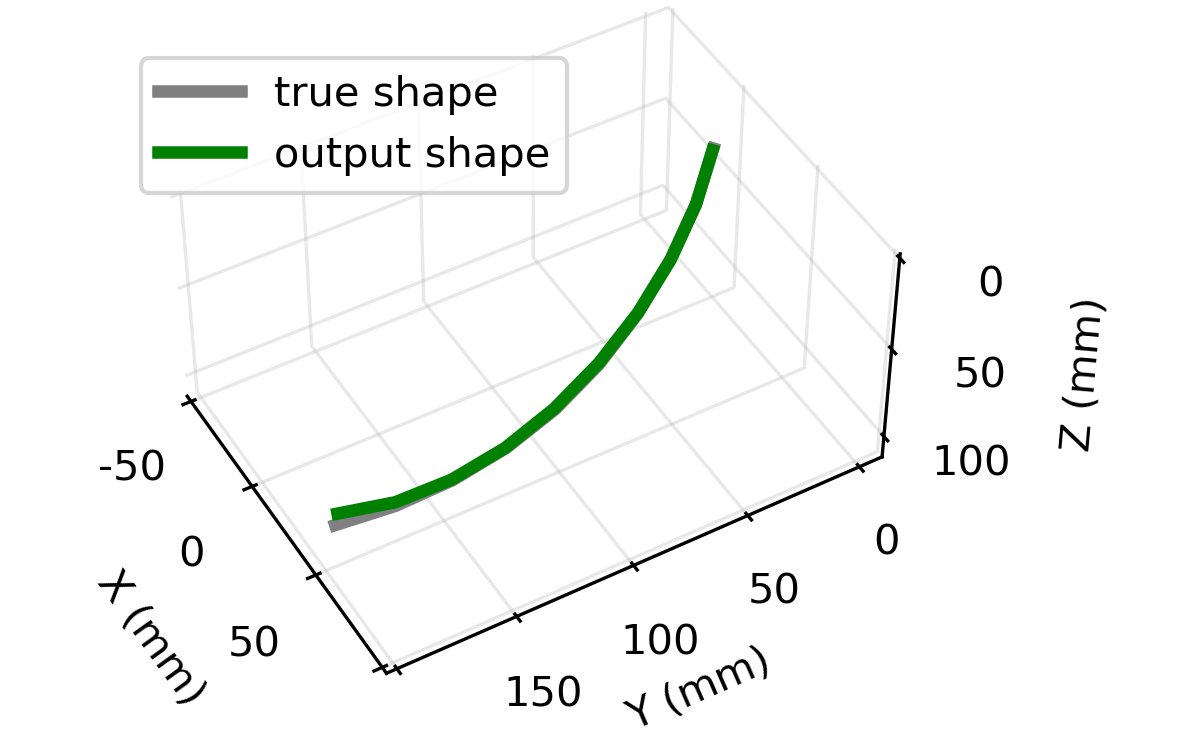}\\
input image  & output shape  \\[6pt]
\end{tabular}
\begin{tabular}{cccc}
\includegraphics[width=.30\linewidth,height=.30\linewidth]{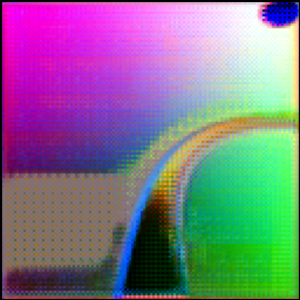} & 
\includegraphics[width=.30\linewidth,height=.30\linewidth]{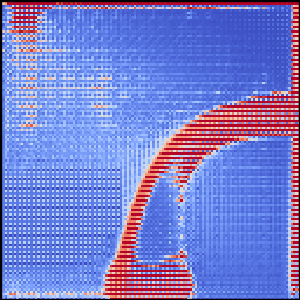} & 
\includegraphics[width=.30\linewidth,height=.30\linewidth]{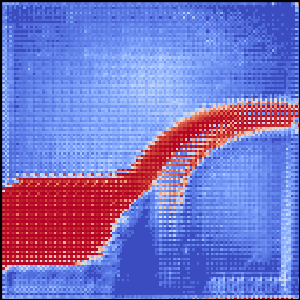}\\
\makecell{centerline\\decoder}  & \makecell{arclength\\decoder} & \makecell{importance\\decoder}  \\[6pt]
\end{tabular}
\caption{Qualitative results on MoSS-SimWide demonstrates the method is robust against changes in camera intrinsic and extrinsic parameters.}
\label{fig:quali_app}
\end{figure}

Quantitative and qualitative results are shown in Table \ref{tab:quant_app} and Fig. \ref{fig:quali_app}, respectively. The mean error of robot shape achieved is slightly higher compared to the MoSS-Sim dataset, which is possibly caused by the smaller dataset size and a higher level of robot self-occlusion. They demonstrate that our approach is robust to different camera intrinsic and extrinsic parameters and can potentially be applied to different imaging modalities depending on the specific application (e.g. laparoscopic imaging and X-ray imaging). 

\section{Conclusion}
\label{sec:conclusion}

We propose a novel monocular shape sensing method for continuum robots, called MoSSNet. Simulated and real datasets collected on a two-segment TDCR demonstrate the method is accurate (mean shape error of $\SI{0.91}{\milli\meter} (0.36\%)$), real-time ($70$ fps), and generalizes to unseen data. The method is also optimized end-to-end and does not require fiducial markers, segmentation, or camera calibration. MoSSNet outperforms existing stereo-vision-based shape sensing methods in terms of real-time capability and has much lower hardware complexity compared to embedded sensing methods. We believe that these promising results present a potential new alternative for continuum robot shape sensing. Additionly, we provide our code and dataset as part of the \href{www.opencontinuumrobotics.ca}{OpenCR Project}, serving as a research benchmarking tool. 

Future work could expand the dataset by incorporating application-specific backgrounds, imaging modalities, and undesired artifacts. This extension would facilitate the development of vision-based methods to address realistic scenarios.
The method could also be extended to incorporate temporal information for improved accuracy and robustness and inspire future work in sensor fusion and shape control.

\section*{Conflict of Interest}
E. L. declares that the research was conducted in the absence of any commercial or financial relationships that could be construed as a potential conflict of interest.

\section*{Acknowledgment}
The authors thank Raquel Urtasun for the insightful discussions related to Section \ref{sec:method} of this paper.


\bibliographystyle{ieeetr}
\bibliography{references}

\end{document}